\def\year{2019}\relax
\documentclass[letterpaper]{article} 
\usepackage{aaai19}  
\usepackage{times}  
\usepackage{helvet}  
\usepackage{courier}  
\usepackage{url}  
\usepackage{graphicx}  

\usepackage{latexsym}
\usepackage{amsmath}
\usepackage{amssymb}
\usepackage{xcolor}
\usepackage{soul}
\usepackage[utf8]{inputenc}
\usepackage{multirow}
\usepackage{makecell}

\begin{document}
%
\title{Adversarial Training for Community Question Answer Selection \\ Based on Multi-scale Matching}

\author{Xiao Yang$^{\dag}$\thanks{Work done at Microsoft.}, Madian Khabsa$^{\P}$\footnotemark[1], Miaosen Wang$^{\S}$\footnotemark[1], Wei Wang$^{\ddag}$ \\ \large{\textbf{Ahmed Hassan Awadallah$^{\ddag}$, Daniel Kifer$^{\dag}$, Lee Giles$^{\dag}$}} \\
$^{\dag}$Pennsylvania State University, $^{\P}$Apple, $^{\S}$Google,  $^{\ddag}$Microsoft \\
xuy111@psu.edu, madian@apple.com, miaosen@google.com, wawe@microsoft.com \\ hassanam@microsoft.com, duk17@psu.edu, giles@ist.psu.edu
}

\maketitle

\begin{abstract}
Community-based question answering (CQA) websites represent an  important source of information. As a result, the problem of matching the most valuable answers to their corresponding questions has become an increasingly popular research topic. We frame this task as a binary (relevant/irrelevant) classification problem, and present an adversarial training framework to alleviate label imbalance issue. We employ a generative model to iteratively sample a subset of challenging negative samples to fool our classification model. Both models are alternatively optimized using REINFORCE algorithm. The proposed method is completely different from previous ones, where negative samples in training set are directly used or uniformly down-sampled. Further, we propose using Multi-scale Matching which explicitly inspects the correlation between words and ngrams of different levels of granularity. We evaluate the proposed method on SemEval 2016 and SemEval 2017 datasets and achieves state-of-the-art or similar performance. 
\end{abstract}

\section{Introduction}
Community-based question answering (CQA) websites such as Yahoo Answer and Quora are important sources of knowledge. They allow users to submit questions and answers covering a wide range of topics. These websites often organize such user-generated content in the form of a question followed by a list of candidate answers. Over time, a large amount of crowd-sourced question/answer pairs has been accumulated, which can be leveraged to automatically answer a newly submitted question.

To fully make use of the knowledge stored in CQA systems, the CQA selection task has received much attention recently. CQA selection task aims at automatically retrieving archived answers that are relevant to a newly submitted question. Since many users tend to submit new questions rather than searching existing questions~\cite{nakov2016takes}, a large number of questions reoccur and they may already be answered by previous content. Several challenges exist for this task, among which \textit{lexical gap} is a fundamental one that differentiate this task from other general-purpose information retrieval (IR) tasks. The term \textit{lexical gap} describes the phenomenon that text in questions may lead to causally related content in answers, rather than semantically related content such as synonymy. In other words, a sentence sharing many overlapping words with the question may not necessarily be a relevant answer.

To tackle the lexical gap problem, many methods explicitly model the correlations between text fragments in questions and answers, and frame this task as a binary classification problem. Under this setting, each question/answer pair is labeled as either relevant or irrelevant. Consequently, the CQA selection task can be approached by first predicting the relevance/confidence score of a candidate answer to a question, then re-ranking these answers to find the most appropriate one.

When treating CQA selection as a binary classification task, a practical issue is label imbalance. First, the total number of relevant answers is naturally smaller than that of irrelevant ones. For example, in SemEval 2016 dataset, only 9.3\% of the total answers are relevant. In SemEval 2017 dataset, the number is even smaller: 2.8\%. Furthermore, in order to better utilize labeled data and provide more training question/answer pairs, many researches augment negative samples by randomly coupling a question with an answer from a \textit{different} question thread. The underlying assumption is that answers from other questions are unlikely to be qualified as relevant to the current question. While such data augmentation can provide much more training samples, it also amplifies the problem of label imbalance. Inspired by Generative Adversarial Nets (GANs)~\cite{goodfellow2014generative}, we present an adversarial training strategy, which employs a generative model $G$ to select a small subset of \textit{challenging} negative samples. A fair amount of negative samples (especially those generated by random coupling described above) can be easily classified (e.g. the topics of answers are completely irrelevant to the questions), making little contribution to the gradients. On the contrary, samples selected by a generative model are expected to be more challenging, therefore are more likely to fool the classification model $D$, and consequently result in larger gradients. By alternately optimizing the generative model and the classification model, we can finally obtain a more robust and accurate classifier.

For the classification model $D$, the ``matching-aggregating'' framework~ \cite{wang2016compare,parikh2016decomposable,zhang2017attentive,wang2017bilateral} is a representative work for CQA task. It first represents each word by embeddings, then exhaustively compares each word in questions to another word in answers. The comparison results are later aggregated by a feed-forward neural network to make final predictions. Various strategies have been proposed for aggregating comparisons, such as max-pooling method~\cite{zhang2017attentive}, attention method~\cite{parikh2016decomposable}, or a combination of various strategies~\cite{wang2017bilateral}. It is shown that such ``matching-aggregating'' framework outperforms Long Short Term Memory (LSTM)~\cite{hochreiter1997long} based methods~\cite{wang2016compare,zhang2017attentive}. Our work also follows the ``matching-aggregating'' framework. However, in addition to word-to-word comparisons, we also explicitly consider comparisons between words and ngrams of different length. The rationale behind is that the semantic meaning of a text fragment is not the simple combination of the meanings of individual words~\cite{stubbs2001words}. By explicitly considering word-to-ngrams comparisons, our model is enforced to capture semantic information at different levels of granularity, and utilize it to assist classification. To obtain word-to-ngrams comparisons, we employ a deep convolutional neural network (CNN) to learn a hierarchical representation for each sentence. Neurons at higher levels compress information of larger context. Representations at different levels of one sentence are then compared with those from the other sentence.

Our contributions are summarized as follows:
\begin{itemize}
	\item We present an adversarial training strategy which employs a generative model to produce challenging negative samples. By alternately optimizing the generative model and the classification model, we are able to significantly improve performance on CQA task.
	\item We extend the current matching-aggregating framework for CQA selection task by also considering matchings from multiple levels of granularity. Experiments show that such multi-scale matching consistently improves performance.
	\item The proposed model ranks first on SemEval 2017 dataset and ranks second on SemEval 2016 dataset (first among methods that do not use external meta information such as answers position in a thread or author's personal information).
\end{itemize}

\section{Related Work}
\subsection{Community Question Answering}
For an automatic community question answer selection system, two main tasks exist: (1) retrieving related questions to a newly submitted question~\cite{jeon2005finding,xue2008retrieval,cai2011learning,zhou2012classification}; and (2) retrieving potentially relevant answers to a newly submitted question~\cite{surdeanu2008learning,lu2013deep,shen2015question,shen2017word,wang2017bilateral,nakov2016takes,zhang2017attentive}. Successfully accomplishing the first task can assist the second task. However, it is not a must step. Techniques for CQA selection can be broadly categorized into three classes: (1) statistical translation models; (2) latent variable models; and (3) deep learning models.

Early work spent great efforts on statistical translation models, which take parallel corpora as input and learn correlations between words and phrase from one corpus and another. For example, \cite{jeon2005finding,xue2008retrieval} use IBM translation model 1 to learn translation probability between question and answer words. Later work has improved upon them by considering phrase-level correlations~\cite{cai2011learning} and entity-level correlations~\cite{singh2012entity}. The proposed Multi-scale Matching model shares similar idea by incorporating word-to-ngrams comparisons, however such comparisons are modeled by a neural network rather than translation probability matrix.

Another line of work explores using topic models for addressing this task. Such approaches~\cite{cai2011learning,ji2012question,shen2015question} usually learn the latent topics of questions and answers, under the assumption that a relevant answer should share a similar topic distribution to the question. Recently, these approaches have been combined with word embeddings~\cite{le2014distributed,shen2015question,zhou2015learning} and translation models~\cite{deepak2017latent}, which have led to further improvements of performance.

With the recent success of deep learning models in multiple natural language processing (NLP) tasks, researchers started to explore deep models for CQA. \cite{nakov2016takes} proposed a feed-forward neural network to predict the pairwise ranking of two candidate answers. \cite{zhou2016learning} trained two auto-encoders for questions and answers respectively which share the intermediate semantic representation. Recently, a number of work has framed this task as a text classification problem, and proposed several deep neural network based models. For example, \cite{tan2015lstm} first encode sentences into sentence embeddings using LSTM, then predict the relationship between questions and answers based on the learned embeddings. However, such approaches ignore direct interactions between words in sentences, therefore their performances are usually limited. Later, \cite{wang2016compare,parikh2016decomposable,zhang2017attentive} proposed a matching-aggregating framework which first exhaustively compares words from one sentence to another, then aggregates the comparison results to make final predictions. Different aggregating strategies have been proposed, such as attentive method~\cite{parikh2016decomposable}, max-pooling method~\cite{zhang2017attentive}, or a combination of various strategies~\cite{wang2017bilateral}. The proposed Multi-scale Matching model also follows such framework, however we explicitly examine comparisons at multiple levels of granularity.

\subsection{Generative Adversarial Nets and NLP}
Generative Adversarial Nets (GANs)~\cite{goodfellow2014generative} was first proposed for generating samples from a continuous space such as images. It consists of a generative model $G$ and a discriminative model $D$. $G$ aims to fit the real data distribution and attempts to map a random noise (e.g. a random sample from a Gaussian distribution) to a real sample (e.g. an image). In contrary, $D$ attempts to differentiate real samples from fake ones generated by $G$. During training, $G$ and $D$ are alternately optimized, forming a mini-max game. A number of extensions to GAN have been proposed to achieve stable training and better visualization results for image generation.
 
The idea of adversarial training can also be applied to NLP tasks. Although such tasks often involve discrete sampling process which is not differentiable, researchers have proposed several solutions such as policy gradient~\cite{sutton2000policy,yu2017seqgan,li2016deep} and Gumbel Softmax trick~\cite{jang2016categorical}. \cite{yu2017seqgan} proposed SeqGAN to generate sequence of words from noises. \cite{li2016deep} adopted adversarial training to improve the robustness of a dialog generation system. A more relevant work to our method is IRGAN~\cite{wang2017irgan}, which applied adversarial training to multiple information retrieval tasks. However, IRGAN models the relationship between two documents solely based on the cosine similarity between two learned sentence embeddings, ignoring all direct interactions between words. In contrary, we explicitly explore comparisons at multiple levels of granularity, and use the aggregated comparison results to measure the relevance.

\section{Method}
In this section, we first formally define the task of community question answer selection by framing it as a binary classification problem, then present details about how to fit this problem in an adversarial training framework. Finally, we describe how to instantiate the generative model and classification model using a Multi-scale Matching Model.

Let $Q = (q_1, q_2, \cdots, q_m)$ and $A = (a_1, a_2, \cdots, a_n)$ be the input question and answer sequence of length $m$ and $n$, respectively. Let $f_{\theta} (Q, A)$ be a score function parameterized by $\theta$ that estimates the relevance between $Q$ and $A$. A higher $f_{\theta} (Q, A)$ value means that an answer is more relevant to the question. Given a question $Q$, its corresponding candidate answer set $\mathbf{A} = \{A_i\}$ can be ranked based on the predicted relevance score. The top ranked answers will be selected as the correct answers. Therefore the answer selection task can be accomplished by solving a binary classification problem.

\subsection{Adversarial Training for Answer Selection}
\textbf{Generative Adversarial Nets} 
Generative Adversarial Nets were first proposed by \cite{goodfellow2014generative}. They consists of two ``adversarial'' models: a generative model (generator $G$) aiming at capturing real data distribution $\text{p}_{data}(\mathbf{x})$, and a discriminative model (discriminator $D$) that estimates the probability that a sample comes from the real training data rather than the generator. Both the generator and discriminator can be implemented by non-linear mapping functions, such as feed-forward neural networks.

The discriminator is optimized to maximize the probability of assigning the correct labels to either training samples or the generated ones. On the other hand, the generator is optimized to maximize the probability of $D$ making a mistake, or equivalently to minimize $\log(1 - D(G(\mathbf{x}))$. Therefore, the overall objective can be summarized as:
\begin{align}\label{eq:gan}
	J(G, D) &= \min_G \max_D \mathbb{E}_{\mathbf{x}\sim p_{data}(\mathbf{x})}[\log D(\mathbf{x})] + \nonumber \\ 
	&\mathbb{E}_{\mathbf{x'}\sim p_G(\mathbf{x}')}[\log(1-D(\mathbf{x}'))]
\end{align}
where the generative model $G$ is written as $p_G(\mathbf{x'})$. During training, we alternately minimize and maximize the same objective function to learn the generator $G$ and the discriminator $D$, respectively.

\textbf{Adversarial Training for Answer Selection} Here we propose an adversarial training framework which uses a Multi-scale Matching model (described in next section) to generate (sample) challenging negative samples and another Multi-scale Matching model to differentiate positive samples from negative ones. In parallel to terminologies used in GANs literature, we will call these two models generator $G$ and discriminator $D$ respectively.

In the context of answer selection, the generator aims to capture real data distribution $p_{data} (A | Q)$ and generate (sample) relevant answers conditioned on the question sentence $Q$. In contrary, the discriminator attempts to distinguish between relevant and irrelevant answers depending on $Q$. Formally, objective function in Equation~\ref{eq:gan} can be rewritten as:
\begin{align}
	J(G, D) &= \min_G \max_D \mathbb{E}_{A\sim p_{data}(A | Q)}[\log D(A | Q)] + \nonumber \\ 
	&\mathbb{E}_{A'\sim p_G(A' | Q)}[\log(1-D(A' | Q))]
\end{align}

We now describe how to build our discriminator and generator using the proposed Multi-scale Matching model. Since our Multi-scale Matching model can be seen as a score function, which measures how relevant an answer $A$ is to a question $Q$, we can directly feed the relevance into a sigmoid function to build our discriminative model. The generator attempts to fit the underlying real data distribution, and based on that, samples answers from the whole answer set in order to fool the discriminator. In order to model this process, we employ another Multi-scale Matching model as a score function and evaluate it on \textit{every} candidate answer. Afterwards, answers with high relevance scores will be sampled with high probabilities. In other words, we would like to sample negative answers from the whole set which is more relevant to $Q$.

Formally, given a set of candidate answers $\mathbf{A} = \{A_i\}$ of a specific question $Q$, the discriminative model $D(A | Q)$ and the generative model $p_G(A | Q)$ is modeled by:
\begin{align}\label{eq:sample}
D(A | Q) &= \sigma (f_{\theta} (Q, A)) \\
p_G(A_i | Q) &= \frac{\exp (f_{\theta'} (Q, A_i))}{\sum_j \exp (f_{\theta'} (Q, A_j))}
\end{align}
with $\sigma$ being a sigmoid function.

Ideally, the score function $f_{\theta'}(Q, A)$ needs to be evaluated on each possible answer. However, the actual size of such answer set can be very large, making such approach computationally infeasible. To address this issue, in practice, for each question we first uniformly sample an alternative answer set $\tilde{\mathbf{A}}$ whose size is much smaller (e.g. 100), then evaluate $f_{\theta'}(Q, A)$ on every answer in set $\tilde{\mathbf{A}}$ and sample top 10 answers. Set $\tilde{\mathbf{A}}$ is constituted by answers from two sources: (1) labeled negative answers for question $Q$; and (2) answers from other questions $\tilde{Q} \neq Q$. Since irrelevant answers are far more than relevant answers, the resulting set $\tilde{\mathbf{A}}$ is unlikely to contain any false negatives.

The original GANs require that both the generator and the discriminator are fully differentiable, so that a gradient-based optimization algorithm can be applied. However, this is not true in our case due to the random sampling step involved in the generator. A number of approaches have been proposed to tackle this problem, such as policy gradient~\cite{sutton2000policy,yu2017seqgan,li2016deep} and Gumbel Softmax trick~\cite{jang2016categorical}. Here we adopt the policy gradient approach. As can be seen in Equation~\ref{eq:gan}, the objective function for optimizing $G$ is expressed as minimizing the expectation of a function evaluated on samples from a probability distribution. Therefore, using the REINFORCE~\cite{sutton2000policy} algorithm, the gradient of $J$ with respect to $G$'s parameters $\theta'$ can be derived as:
\begin{align}
&\nabla_{\theta'} J(G, D) = \nabla_{\theta'} \mathbb{E}_{A'\sim p_G(A' | Q)}[\log(1-D(A' | Q))] \nonumber \\
	&= \sum_{A' \in \mathbf{A}} \nabla_{\theta'} p_G(A' | Q) \log(1-D(A' | Q)) \nonumber \\
	&= \mathbb{E}_{A'\sim p_G(A' | Q)} [\nabla_{\theta'} \log p_G(A' | Q) \log(1-D(A' | Q))] \nonumber \\
	&\simeq \frac{1}{|\tilde{\mathbf{A}}|} \sum_{A' \in \tilde{\mathbf{A}}} \nabla_{\theta'} \log p_G(A' | Q) \log(1-D(A' | Q))
\end{align}
where in the last step the expectation is approximated by sampling. The term $\log(1-D(A' | Q)$ can be seen as the received reward when a policy $p_G$ takes an action of choosing answer $A'$. In practice we also use the averaged reward from last epoch as a reward baseline.

\subsection{Multi-scale Matching Model}\label{multiscale}
\begin{figure}
\centering
\includegraphics[width=1.0\linewidth]{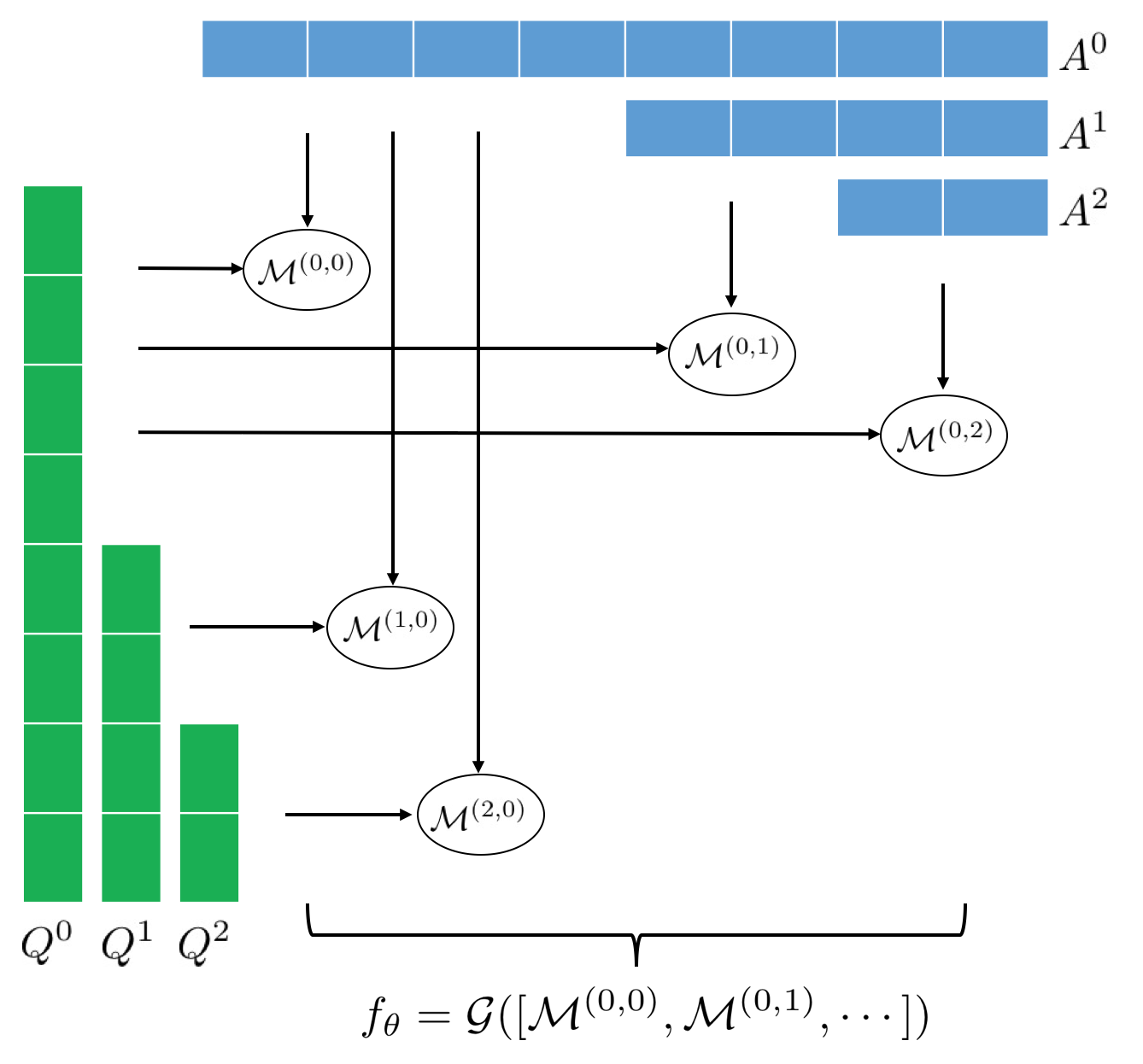}
\vspace{-1em}
\caption{Architecture of the proposed Multi-scale Matching model. The model first learns a hierarchy of representations for each sentence, then applies matching function $\mathcal{M}$ between representations at multiple levels of granularity. The matching results are finally aggregated to predict the relevance score $f_\theta$.}
\vspace{-0.5em}
\label{figure:architecture}
\end{figure}

In this section we describe the details of our Multi-scale Matching model. The goal of the proposed Multi-scale Matching Model is to estimate a relevance score of a question/answer pair. Our model follows the ``matching-aggregating'' framework. Different from \cite{parikh2016decomposable} which only consider word-to-word matches and \cite{wang2017bilateral} which implicitly consider word-to-ngrams matches using an attention model, we explicitly examine matches between words and ngrams of different lengths. In this way, the proposed model is enforced to leverage context information at different levels of granularity. The architecture is illustrated in Figure~\ref{figure:architecture}.

\textbf{Word and Ngram Embeddings} For either a question or answer sentence, we represent each word with a $d$-dimensional real-valued vector. In this work, we use pre-trained word embeddings from GloVe~\cite{pennington2014glove}, which has shown its effectiveness in multiple natural language processing tasks. For each sentence, our model learns a hierarchy of representations using a convolutional neural network. Formally, for a sentence $Q = (q_1, q_2, \cdots, q_m)$ where each word $q_i$ is represented by the corresponding word embeddings, a series of convolution blocks are applied:
\begin{align}
	Q^k = \text{conv\_block}^k(Q^{k-1}) \qquad
	Q^0 = Q
\end{align}

Here $Q^k$ is the resulting feature maps after applying the $k$-th convolution block. A convolution block consists of a convolution layer, followed by a batch normalization~\cite{ioffe2015batch} layer, a rectified linear unit~\cite{glorot2011deep} layer and a max pooling layer. The kernel size of the convolution layer is 3, the stride is 1 and the number of output channels is 128. At the end, a hierarchy of feature representations $(Q^0, Q^1, \cdots, Q^K)$ is learned, which compresses the semantic information at different levels of granularities. For example, $Q_i^0$ represents the $i$-th word embedding, while $Q_i^1$ represents the context information from a 5-gram centered at $i$, since the receptive field is 5.

Similar process is applied to the answer sentence $A = (a_1, a_2, \cdots, a_n)$, resulting in another hierarchy of feature representations $(A^0, A^1, \cdots, A^K)$.

\textbf{Multi-scale Matching and Aggregating} For a specific pair of feature representations $Q^u$ and $A^v$, we can define a matching function $\mathcal{M} (Q^u, A^v)$ to measure the relation between them. The final score $f_{\theta} (Q, A)$ can be calculated by aggregating $\{\mathcal{M} (Q^u, A^v)\}$, abbreviated as $\{\mathcal{M}^{(u, v)}\}$.

\begin{figure}
\centering
\includegraphics[width=1.0\linewidth]{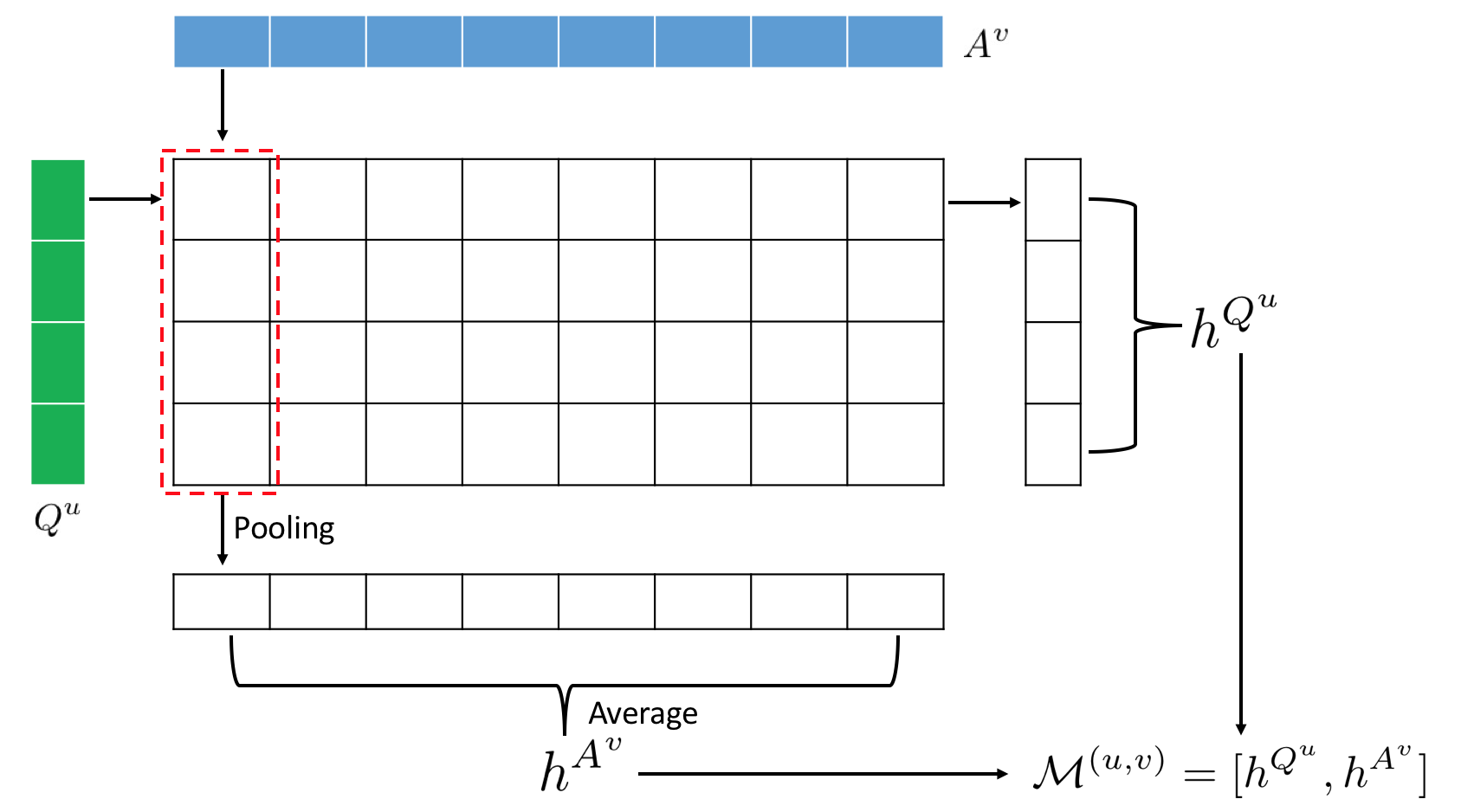}
\vspace{-1em}
\caption{Illustration of the matching between two representations.}
\vspace{-1em}
\label{figure:matching}
\end{figure}

Multiple ways exist on how to realize the matching function $\mathcal{M}$, for example the max-pooling matching~\cite{zhang2017attentive} and attentive matching~\cite{parikh2016decomposable}. Here we adopt the max-pooling matching method due to its simplicity. First, we compare each time-step in $Q^u$ and $A^v$ using a non-linear function $\mathcal{H}$: 
\begin{align}
	h_{(i, j)} = \mathcal{H}([Q_i^u, A_j^v])
\end{align}
where the function $\mathcal{H}$ is implemented by a two-layer feed-forward neural network and $[\cdot]$ denotes concatenation. For each time-step $i$ in $Q^u$, we aggregate comparison results by element-wise max-pooling and obtain a single vector $h_{(i, \cdot)}$:
\begin{align}
	h_{(i, \cdot)} = \text{Pooling} (h_{(i, 1)}, h_{(i, 2)}, \cdots, h_{(i, n)})
\end{align}
Figure~\ref{figure:matching} shows a diagram of such aggregation process.

Similarly, for each time-step $j$ in $A^v$, we can also aggregate the comparison results and obtain:
\begin{align}
	h_{(\cdot, j)} = \text{Pooling} (h_{(1, j)}, h_{(2, j)}, \cdots, h_{(m, j)})
\end{align}

Now we have two sets of comparison vectors $\{h_{(i, \cdot)}\}$ and $\{h_{(\cdot, j)}\}$, we can aggregate over each set by averaging:
\begin{align}
	h^{Q^u} = \frac{1}{m}\sum_i^{m} h_{(i, \cdot)} \qquad
	h^{A^v} = \frac{1}{n}\sum_j^{n} h_{(\cdot, j)}
\end{align}
then formulate the matching function $\mathcal{M}$ as the concatenation of $h^{Q^u}$ and $h^{A^v}$:
\begin{align}
	\mathcal{M}^{(u, v)} = [h^{Q^u}, h^{A^v}]
\end{align}

Based on the defined the matching function $\mathcal{M}$ above, the score function $f_{\theta} (Q, A)$ can be formulated as:
\begin{align}\label{eq:score0}
	f_{\theta} (Q, A) = \mathcal{G} ([\{\mathcal{M}^{(u, v)}\}])
\end{align}
where $[\{\mathcal{M}^{(u, v)}\}]$ denotes the concatenation of all possible matching results $\mathcal{M}^{(u, v)}$ for $u=0, 1, \cdots, K$ and $v = 0, 1, \cdots, K$, and $\mathcal{G}$ is a real-value function implemented by a two-layer feed-forward neural network. Equation~\ref{eq:score0} indicates that all possible word-to-word, word-to-ngram and ngram-to-ngram matches are being considered. A simpler way is to formulate the score function as:
\begin{align}\label{eq:score1}
	f_{\theta}  (Q, A) = \mathcal{G} ([\{\mathcal{M}^{(0, v)}\}, \{\mathcal{M}^{(u, 0)}\}])
\end{align}
meaning that we only consider word-to-word and word-to-ngram matches, and ignore all ngram-to-ngram matches. It is also clear that the way in \cite{wang2016compare,zhang2017attentive,parikh2016decomposable} is equivalent to formulating the score function as:
\begin{align}\label{eq:score2}
	f_{\theta} (Q, A) = \mathcal{G} (\mathcal{M}^{(0, 0)})
\end{align}
indicating that only word-to-word matching is considered. The time complexity of these three ways are $O(K^2)$, $O(K)$ and $O(1)$ respectively if we consider each $\mathcal{M}^{(u, v)}$ as an $O(1)$ operation.

In this work, we adopt the second way as described in Equation~\ref{eq:score1}, since the first way (Equation~\ref{eq:score0}) is computationally expensive and the third way (Equation~\ref{eq:score2}) may not fully utilize context information conveyed in ngrams. Results in Section~\ref{sec:experiments} shows that the second way leads to better performance compared with the third way.

\section{Experiments}\label{sec:experiments}
In this section, we evaluate the proposed method on two benchmark datasets: SemEval 2016 and SemEval 2017. Ablation experiments are conducted on both datasets to demonstrate the effectiveness of the proposed adversarial training strategy and Multi-scale Matching model.

\subsection{Datasets and Evaluation}
\textbf{Datasets} SemEval 2016~\cite{nakov2016semeval} and SemEval 2017~\cite{nakov2017semeval} datasets are used for Task 3, Subtask C (Question-External Comment Similarity) of SemEval 2016 and SemEval 2017 challenge, respectively. SemEval 2016 contains 387 original questions from Qatar Living website among which 70 questions are used as test set. Each question is associated with the top 10 related questions (retrieved by a search engine) and their corresponding top 10 answers appearing in the thread. As a result, each question is associated with 100 candidate answers, and the ultimate goal is to \textit{re-rank} these 100 answers according to their relevance to the original question. SemEval 2017 is the most recent dataset where additional 80 questions are used as test set. This is a more challenging dataset compared with SemEval 2016, since it has a much more imbalanced label distribution: only 2.8\% candidate answers are labeled as relevant, whereas in SemEval 2016 the number is 9.3\%. 

\textbf{Evaluation Metrics} We use the official evaluation measure for the competition which is mean average precision (MAP) calculated over the top 10 ranked answers. We also report mean reciprocal rank (MRR), which is another widely-used information retrieval measure. 

\subsection{Training Hyper-parameters}
The model weights are optimized using Adam~\cite{kingma2014adam} optimization method. The initial learning rate is $1e-4$ and is decayed by 5 for every 10 epochs. We use L2 regularization on model weights with a coefficient of $1e-6$ and a drop out rate of 0.2.

\subsection{Results on SemEval 2017}
Table~\ref{table:semeval} summarizes the results of different methods on SemEval 2017 dataset. For our methods, the term ``single'' denotes that we only consider word-to-word matches as in Equation~\ref{eq:score2}, while ``multi'' means that we consider both word-to-word and word-to-ngrams matches as in Equation~\ref{eq:score1}. The term ``adversarial'' means that we employ an additional generative model to produce challenging adversarial samples to fool the discriminative model during training. From the table we can see that using Multi-scale Matching consistently improves the performance. With only a discriminative model, MAP is increased from 14.67 to 14.80. With adversarial training, MAP is increased from 17.25 to 17.91.

\begin{table}
\begin{center}
\begin{tabular}{| p{15em} | c c |} \hline
	Method & MAP & MRR \\ \hline
    Baseline (IR) & 9.18 & 10.11 \\
    Baseline (random) & 5.77 & 7.69 \\ \hline
	\cite{tian2017ecnu} & 10.64 & 11.09 \\
	\cite{zhang2017furongwang} & 13.23 & 14.27 \\
	\cite{xie2017eica} & 13.48 & 16.04 \\
	\cite{filice2017kelp} & 14.35 & 16.07 \\
	\cite{koreeda2017bunji} & 14.71 & 16.48 \\
	\cite{nandi2017iit} & 15.46 & 18.14 \\
    Contrs.~\cite{koreeda2017bunji} & 16.57 & 17.04 \\ \hline
	Ours (single) & 14.67 & 16.75\\
	Ours (multi) & 14.80 & 17.57\\
	Ours (single+adversarial, D) & 17.25 & 17.62\\
	Ours (multi+adversarial, D) & \textbf{17.91} & \textbf{18.64}\\ \hline
    Ours (single+adversarial, G) & 13.31 & 15.07\\
    Ours (multi+adversarial, G) & 14.33 & 16.51\\ \hline
\end{tabular}
\end{center}
\vspace{-1em}
\caption{Performance on SemEval 2017 dataset. ``Contrs'' denotes non-primary submission.}
\vspace{-0.5em}
\label{table:semeval}
\end{table}

With adversarial training, both our single-scale and multi-scale models are significantly improved and outperform previous methods which are primarily based on feature engineering~\cite{filice2017kelp,xie2017eica,nandi2017iit} and neural networks~\cite{tian2017ecnu,zhang2017furongwang,koreeda2017bunji}. For single-scale model, the MAP is increased from 14.67 to 17.25, while for multi-scale model, the number is increased from 14.80 to 17.91. This demonstrates the effectiveness of utilizing a generative model to produce challenging negative samples. Since in SemEval 2017 dataset each question is associated with 100 candidate answers and only 2.8\% of them are labeled as relevant, the class labels are severely imbalanced. By having a generative model, we are able to select more challenging adversarial samples as training proceeds, resulting in a more robust discriminator. Table~\ref{table:semeval} also shows the performance achieved by generators that are checkpointed at the same stage as the discriminators. The MAP is 13.31 and 14.33 for single-scale and multi-scale model respectively, significantly outperforming a random baseline and a information-retrieval (IR) based baseline. This shows that our generators also learned useful information. Consequently, the negative samples selected by the generators are much more challenging that those selected randomly or by a IR based approach.

\subsection{Results on SemEval 2016}
Table~\ref{table:semeval2016} summarizes our results on SemEval 2016 dataset (numbers are extracted from \cite{nakov2016semeval}). Using Multi-scale Matching model boosts MAP for both direct and adversarial training, and the improvements are more significant than those on SemEval 2017. With adversarial training, our single-scale and multi-scale models are significantly improved, achieving a MAP of 52.09 and 53.38, respectively. Again we list the performances achieved by the generators. It can be seen that the generators significantly outperform random baseline but is slightly worse than IR and chronological ranking based baseline.

However, when comparing with other prior methods, our method ranks only second in terms of MAP among primary submissions. We hypothesize that this is because SemEval 2016 dataset is more balanced. In the test set of SemEval 2016, 9.3\% of the candidate answers are labeled as relevant, which is roughly 3.3 times more balanced than that of SemEval 2017. This makes improvements made by adversarial training less significant. The motivation of adversarial training is to let a generator down-sample negative examples in a smarter way, so that as training proceeds, more and more challenging negative examples can be used to train the discriminator. If labels are already balanced, then directly training a discriminator is likely to yield good results. Note that our result is still an important achievement given that many other methods, including the best performing method, make use of meta information (e.g. answers' positions in threads; whether an answer is written by the author of the question; whether the author of an answer is active in the thread), while our method only relies on textual information.

\begin{table}
\begin{center}
\begin{tabular}{| l | c c |} \hline
	Method & MAP & MRR \\ \hline
    Baseline (IR+chronological) & 40.36 & 45.83 \\
    Baseline (random) & 15.01 & 15.19 \\ \hline
	\cite{marc2016uhprhlt} & 43.20 & 47.79 \\
	\cite{wu2016ecnu} & 46.47 & 51.41 \\
	\cite{barron2016convkn} & 47.15 & 51.43 \\
    \cite{mihaylov2016semanticz} & 51.68 & 55.96 \\
	\cite{filice2016kelp} & 52.95 & 59.23 \\
	\cite{mihaylova2016super} & 55.41 & \textbf{61.48} \\
    Contrs~\cite{filice2016kelp} & \textbf{55.58} & 61.19 \\\hline
	Ours (single) & 48.11 & 54.25\\
	Ours (multi) & 49.25 & 54.89\\
	Ours (single+adversarial, D) & 52.09 & 59.64\\
	Ours (multi+adversarial, D) & 53.38 & 60.64\\ \hline
    Ours (single+adversarial, G) & 36.31 & 41.19\\
    Ours (multi+adversarial, G) & 37.14 & 41.84\\ \hline
\end{tabular}
\end{center}
\vspace{-1em}
\caption{Performance on SemEval 2016 dataset.  ``Contrs'' denotes non-primary submission. Note that both \cite{mihaylova2016super} and \cite{filice2016kelp} utilized meta information (e.g. answers' positions in threads; whether an answer is written by the author of the question; whether the author of an answer is active in the thread) while our method only relies on textual information.}
\vspace{-0.5em}
\label{table:semeval2016}
\end{table}

\begin{table*}
\vspace{-1em}
\begin{tabular}[t]{| p{.25\textwidth} | p{.70\textwidth} |} \hline
	Question & Results \\ \hline
    \textbf{Question}: Does anyone know if there is a dog kennel/hotel in Qatar; where we can have someone to look after our dogs... & \parbox[t]{.70\textwidth}{\textbf{Correct}: Ok; I have a cat. But we take him to Pampered Pets whenever we travel. They also board dogs...\\
                         \textbf{Gen.1}: Dogs as pets are not allowed in Islam but if there was a reason such for security; hunting is allowed...\\
                         \textbf{Gen.2}: ... Have you had dogs before? Where will it live? Have you thought about what you will do when you go on vacation? ...\\
                         \textbf{Gen.3}: ... and that the dog be taken out.' This prohibition is limited to keeping dogs without need or benefit.} \\ \hline
	\textbf{Question}: We now feel ready to explore Qatar and was wondering if anyone can suggest a sandy beach that's suitable for young children? & \parbox[t]{.70\textwidth}{\textbf{Correct}: Best places are Fuwarait beach - swim straight out and there are lots of corals; different fish...\\
    \textbf{Gen.1}: Some hotels have beaches. Otherwise; like DaRuDe said; you gotta drive. Get a Marhaba book...\\
    \textbf{Gen.2}: Mostly any beach in the Mediterranean Sea; some in Southeast Asia; South America and Australia...\\
    \textbf{Gen.3}: with a beautiful sandy beach; a breeze from the ocean; a peaceful and restful place to lay your head; a great pool...} \\ \hline
\end{tabular}
\vspace{-1em}
\caption{Example results. For each question, we show the correctly predicted answer by our discriminator, and three top ranked negative examples (irrelevant answers) ranked by our generator.}
\vspace{-0.5em}
\label{table:examples}
\end{table*}

\subsection{Examples}
Table~\ref{table:examples} shows several example outputs of our Multi-scale Matching model. For each question sentence, we show the correctly ranked top answers by our discriminator, and three top ranked negative examples (here irrelevant answers) returned by our generator. Some long sentences may be truncated to highlight relevant parts. It can be seen that the selected negative examples are somewhat related to the questions, therefore the discriminator is fed with more challenging negative examples, compared with random sampling. This process can be viewed as an active way of hard example mining, which boosts the performance of the discriminator.

\section{Conclusions}
We framed the community question answer selection task as a binary classification problem, and presented an adversarial training strategy to alleviate the label imbalance problem. A generative model is introduced to produce challenging negative samples in order to improve the performance of a discriminator. Furthermore, we proposed a Multi-scale Matching model which is enforced to examine context information at different levels of granularities.The proposed method is evaluated on SemEval 2016 and 2017 datasets and achieved state-of-the-art or similar performance. Future work would investigate the stability of GAN training which remains an open research question, especially when discrete sampling is involved.

\noindent \textbf{Acknowledgements}: We gratefully acknowledge partial support from NSF grant CCF 1317560 and a hardware grant from NVIDIA. This work initiated during Xiao Yang's internship at Microsoft.

\bibliography{aaai}
\bibliographystyle{aaai}

\end{document}